\documentclass[11pt]{article}
\usepackage{acl2012}
\usepackage{natbib}
\renewcommand{\cite}[2][]{\citep[#1]{#2}}
\renewcommand{\newcite}[2][]{\citet[#1]{#2}}

\usepackage[footnotesize]{caption}
\usepackage{amsmath,mathrsfs}
\usepackage{times}
\usepackage{url}
\usepackage{latexsym}
\usepackage{amssymb}
\usepackage{amsfonts}
\usepackage{mathtools}
\usepackage{multirow}
\usepackage{hhline}
\usepackage{graphicx}
\usepackage{verbatim}
\usepackage{upgreek}
\usepackage{rotating}
\usepackage[usenames,dvipsnames]{color}
\usepackage{algorithm}
\usepackage[noend]{algpseudocode}
\usepackage{enumitem}
\usepackage{xspace}
\usepackage{tikz}
\usepackage{tikz-dependency}
\usepackage{appendix}
\usepackage{fixltx2e}
\usepackage{subfig}
\usepackage{grffile}
\usepackage{array}
\usepackage[hidelinks]{hyperref}

\newcommand{\deprel}[1]{\texttt{#1}}
\newcommand{\rel}{r}
\usepackage{bm}
\renewcommand{\vec}[1]{{\boldsymbol{\mathbf{#1}}}}
\newcommand{\ul}{\vec{u}}
\newcommand{\p}{p}
\newcommand{\mb}[1]{\mathbf{#1}}
\DeclareMathOperator{\loss}{loss}
\DeclareMathOperator{\cnt}{\text{count}}
\newcommand{\pos}[1]{\texttt{#1}}
\DeclareMathOperator{\ecnt}{\text{ecount}}
\newcommand{\tagset}{\mathcal{T}}
\newcommand{\vs}{\vec{s}}
\newcommand{\defn}[1]{\textbf{#1}}
\newcommand{\bdry}{\texttt{\#}\xspace}
\DeclareMathOperator*{\mean}{mean}
\DeclareMathOperator{\weight}{w}
\newcommand{\RV}{R_{\texttt{V}}}
\newcommand{\RN}{R_{\texttt{N}}}
\DeclareMathOperator{\sigmoid}{sigmoid}
\DeclareMathOperator{\relu}{relu}

\title{Fine-Grained Prediction of Syntactic Typology: \\ Discovering Latent Structure with {\em Supervised} Learning}

\author{Dingquan Wang \and Jason Eisner  \\
        Department of Computer Science, Johns Hopkins University \\
        \texttt{\{wdd,eisner\}@jhu.edu}}
\date{\today{}}

\begin{document}

\maketitle

\begin{abstract}
  We show how to predict the basic word-order facts of a novel language given only a corpus of part-of-speech (POS) sequences.  We predict how often direct objects follow their verbs, how often adjectives follow their nouns, and in general the directionalities of all dependency relations.  Such typological properties could be helpful in grammar induction.  While such a problem is usually regarded as unsupervised learning, our innovation is to treat it as {\em supervised} learning, using a large collection of realistic synthetic languages as training data.  The supervised learner must identify {\em surface} features of a language's POS sequence (hand-engineered or neural features) that correlate with the language's {\em deeper} structure (latent trees).  In the experiment, we show: 1) Given a small set of real languages, it helps to add many synthetic languages to the training data. 2) Our system is robust even when the POS sequences include noise. 3) Our system on this task outperforms a grammar  induction baseline by a large margin.

\end{abstract}

\vspace{-8pt}
\section{Introduction}\label{sec:intro}

Descriptive linguists often characterize a human language by its {\em typological properties}.  For instance, English is an SVO-type language because its basic clause order is Subject-Verb-Object (SVO), and also a prepositional-type language because its adpositions normally precede the noun.  Identifying basic word order must happen early in the acquisition of syntax, and presumably guides the initial interpretation of sentences and the acquisition of a finer-grained grammar.  In this paper, we propose a method for doing this.  While we focus on word order, one could try similar methods for other typological classifications (syntactic, morphological, or phonological).\looseness=-1

The problem is challenging because the language's true word order statistics are computed from syntax trees, whereas our method has access only to a POS-tagged corpus.  Based on these POS sequences alone, we predict the {\em directionality} of each type of dependency relation.  We define the directionality to be a real number in $[0,1]$: the fraction of tokens of this relation that are ``right-directed,'' in the sense that the child (modifier) falls to the right of its parent (head). For example, the \deprel{dobj} relation points from a verb to its direct object (if any), so a directionality of $0.9$---meaning that 90\% of \deprel{dobj} dependencies are right-directed---indicates a dominant verb-object order. (See Table \ref{tb:typo} for more such examples.) Our system is trained to
  predict the relative frequency of rightward dependencies for each of 57 dependency types from the Universal Dependencies project (UD). We assume that all languages draw on the same set of POS tags and dependency relations that is proposed by the UD project (see \S\ref{sec:data}), so that our predictor works across languages.

Why do this? \newcite{liu-2010} has argued for using these directionality numbers in $[0,1]$ as  fine-grained and robust {\em typological descriptors}.  We believe that these directionalities could also be used to help define an {\em initializer, prior, or regularizer} for tasks like grammar induction or syntax-based machine translation.  Finally, the vector of directionalities---or the feature vector that our method extracts in order to predict the directionalities---can be regarded as a {\em language embedding} computed from the POS-tagged corpus.  This language embedding may be useful as an input to multilingual NLP systems, such as the cross-linguistic neural dependency parser of \newcite{ammar2016one}.  In fact, some multilingual NLP systems
already condition on typological properties looked up in the World Atlas of Language Structures, or WALS \cite{wals}, as we review in \S\ref{sec:relatedwork}.  However, WALS does not list all properties of all languages, and may be somewhat inconsistent since it collects work by many linguists.  Our system provides an automatic alternative as well as a methodology for generalizing to new properties.

\begin{table}[t]
\begin{small}
\begin{tabular}{|@{\centering\hspace{.2em}}c@{\hspace{.2em}}|@{\hspace{.0em}}c@{\hspace{0em}}|}
\hline
Typology&Example\\\hline
\parbox{2cm}{Verb-Object\\(English)}&\parbox{4.5cm}{
\begin{dependency}[theme = simple]
   \tikzstyle{every node}=[font=\footnotesize]
    \begin{deptext}[column sep=.05em]
    \texttt{She}\&\texttt{gave}\&\texttt{me}\&\texttt{a}\&\texttt{raise}\\
    \end{deptext}
\depedge[arc angle=20]{2}{5}{dobj}
\end{dependency}
}\\
\parbox{2cm}{Object-Verb\\(Hindi)}&\parbox{5.5cm}{
\begin{dependency}[theme = simple]
   \tikzstyle{every node}=[font=\footnotesize]
    \begin{deptext}[column sep=.05em]
    \texttt{She}\&\texttt{me}\&\texttt{a}\&\texttt{raise}\&\texttt{gave}\\
    \texttt{vah}\&\texttt{mujhe}\&\texttt{ek}\&\texttt{uthaane}\&\texttt{diya}\\
    \end{deptext}
\depedge[arc angle=20]{5}{2}{dobj}
\end{dependency}
}\\\hline
\parbox{2cm}{Prepositional\\(English)}&\parbox{4.5cm}{
\begin{dependency}[theme = simple]
   \tikzstyle{every node}=[font=\footnotesize]
    \begin{deptext}[column sep=.05em]
    \texttt{She}\&\texttt{is}\&\texttt{in}\&\texttt{a}\&\texttt{car}\\
    \end{deptext}
\depedge[arc angle=20]{5}{3}{case}
\end{dependency}
}\\
\parbox{2cm}{Postpositional\\(Hindi)}&\parbox{5.5cm}{
\begin{dependency}[theme = simple]
   \tikzstyle{every node}=[font=\footnotesize]
    \begin{deptext}[column sep=.05em]
    \texttt{She}\&\texttt{a}\&\texttt{car}\&\texttt{in}\&\texttt{is}\\
    \texttt{vah}\&\texttt{ek}\&\texttt{kaar}\&\texttt{mein}\&\texttt{hai}\\
    \end{deptext}
\depedge[arc angle=20]{3}{4}{case}
\end{dependency}
}\\\hline
\parbox{2.3cm}{Adjective-Noun\\(English)}&\parbox{4.5cm}{
\begin{dependency}[theme = simple]
   \tikzstyle{every node}=[font=\footnotesize]
    \begin{deptext}[column sep=.05em]
    \texttt{This}\&\texttt{is}\&\texttt{a}\&\texttt{red}\&\texttt{car}\\
    \end{deptext}
\depedge[arc angle=20]{5}{4}{amod}
\end{dependency}
}\\
\parbox{2.3cm}{Noun-Adjective\\(French)}&\parbox{5.5cm}{
\begin{dependency}[theme = simple]
   \tikzstyle{every node}=[font=\footnotesize]
    \begin{deptext}[column sep=.05em]
    \texttt{This}\&\texttt{is}\&\texttt{a}\&\texttt{car}\&\texttt{red}\\
    \texttt{Ceci}\&\texttt{est}\&\texttt{une}\&\texttt{voiture}\&\texttt{rouge}\\
    \end{deptext}
\depedge[arc angle=20]{4}{5}{amod}
\end{dependency}
}\\\hline
\end{tabular}
\end{small}
\caption{\label{tb:typo}Three typological properties in the World Atlas of Language Structures \cite{wals}, and how they affect the directionality of Universal Dependencies relations.}
\end{table}

More broadly, we are motivated by the challenge of determining the structure of a language from its superficial features.  Principles \& Parameters theory \cite{chomsky_lectures_1981,chomsky-lasnik-1993} famously---if controversially---hypothesized that human babies are born with an evolutionarily tuned system that is specifically adapted to natural language, which can predict typological properties (``parameters'') by spotting telltale configurations in purely linguistic input. Here we investigate whether such a system can be tuned by gradient descent.  It is at least plausible that useful superficial features do exist: e.g., if nouns often precede verbs but rarely follow verbs, then the language may be verb-final.

\section{Approach}

We depart from the traditional approach to latent structure discovery, namely unsupervised learning.
Unsupervised syntax learners in NLP tend to be rather inaccurate---partly because they are failing to maximize an objective that has many local optima, and partly because that objective does not capture all the factors that linguists consider when assigning syntactic structure.
Our remedy in this paper is a supervised approach.  We simply imitate how linguists have analyzed other languages.  This supervised objective goes beyond the log-likelihood of a PCFG-like model given the corpus, because linguists do not merely try to predict the surface corpus.  Their dependency annotations may reflect a cross-linguistic theory that considers semantic interpretability and equivalence, rare but informative phenomena, consistency across languages, a prior across languages, and linguistic conventions (including the choice of latent labels such as \deprel{dobj}).  Our learner does not consider these factors explicitly, but we hope it will identify correlates (e.g., using deep learning) that can make similar predictions.
Being supervised, our objective should also suffer less from local optima.  Indeed, we could even set up our problem with a {\em convex} objective, such as (kernel) logistic regression, to predict each directionality separately.

Why hasn't this been done before?  Our setting presents unusually sparse data for supervised learning, since {\em each training example is an entire language}.  The world presumably does not offer enough natural languages---particularly with machine-readable corpora---to train a good classifier to detect, say, Object-Verb-Subject (OVS) languages, especially given the class imbalance problem that OVS languages are empirically rare, and the non-IID problem that the available OVS languages may be evolutionarily related.\footnote{Properties shared within an OVS language family may appear to be consistently predictive of OVS, but are actually confounds that will not generalize to other families in test data.} We mitigate this issue by training on the Galactic Dependencies treebanks \cite{wang-eisner-2016}, a collection of more than 50,000 human-like synthetic languages.  The treebank of each synthetic language is generated by stochastically permuting the subtrees in a given real treebank to match the word order of other real languages. Thus, we have many synthetic languages that are Object-Verb like Hindi but also Noun-Adjective like French.  We know the true directionality of each synthetic language and we would like our classifier to predict that directionality, just as it would for a real language.  We will show that our system's accuracy benefits from fleshing out the training set in this way, which can be seen as a form of regularization.

A possible criticism of our work is that obtaining the input POS sequences requires human annotators, and perhaps these annotators could have answered the typological classification questions as well.  Indeed, this criticism also applies to most work on grammar induction.  We will show that our system is at least robust to noise in the input POS sequences (\S\ref{sec:robust}).  In the future, we hope to devise similar methods that operate on raw word sequences.

\section{Data}\label{sec:data}

We use two datasets in our experiment:

\paragraph{UD: Universal Dependencies version 1.2 \cite{UNIVDEP-1.2}} A collection of dependency treebanks for 37 languages, annotated in a consistent style with POS tags and dependency relations.

\paragraph{GD: Galactic Dependencies version 1.0 \cite{wang-eisner-2016}} A collection of projective dependency treebanks for 53,428 synthetic languages, using the same format as UD.  The treebank of each synthetic language is generated from the UD treebank of some real language by stochastically permuting the dependents of all nouns and/or verbs to match the dependent orders of other real UD languages.  Using this ``mix-and-match'' procedure, the GD collection fills in gaps in the UD collection---which covers only a few possible human languages.\looseness=-1

\section{Task Formulation}\label{sec:task}
\newcommand{\uc}{u}
\newcommand{\lang}{L}
\newcommand{\udrel}{\mathcal{R}}
\newcommand{\nudrel}{N}
\newcommand{\rrel}{\to}
\newcommand{\lrel}{\leftarrow}
\newcommand{\rrrel}{\overset{\rel}{\rrel}}
\newcommand{\rlrel}{\overset{\;\rel}{\lrel}}
We now formalize the setup of the fine-grained typological prediction task.  Let $\udrel$ be the set of universal relation types, with $\nudrel=|\udrel|$.  We use
  $\rrrel$ to denote a rightward dependency token with label $\rel\in\udrel$.

\textbf{Input} for language $L$: A POS-tagged corpus $\ul$.  (``$\ul$'' stands for ``unparsed.'')

\textbf{Output} for language $L$: Our system predicts $\p(\rrel \mid \rel, \lang)$, the probability that a token in language $\lang$ of an $r$-labeled dependency will be right-oriented.   It predicts this for {\em each} dependency relation type $\rel\in \udrel$, such as $\rel=\deprel{dobj}$.
Thus, the output is a vector of predicted probabilities $\mb{\hat{\p}} \in [0,1]^\nudrel$.

\textbf{Training:} We set the parameters of our system using a collection of training pairs $(\ul, \mb{\p}^*)$, each of which corresponds to some UD or GD training language $\lang$.  Here $\mb{\p}^*$ denotes the true vector of probabilities as empirically estimated from $L$'s treebank.

\textbf{Evaluation}: Over pairs $(\ul,\mb{p}^*)$ that correspond to held-out {\em real} languages, we evaluate the expected loss of the predictions $\mb{\hat{\p}}$.  We use $\varepsilon$-insensitive loss\footnote{Proposed by V.\@ Vapnik for support vector regression.} with $\varepsilon=0.1$, so our evaluation metric is
\begin{align}
  \sum_{\rel \in \udrel}\p^*(\rel \mid \lang)\cdot\loss_\varepsilon(\hat{\p}(\rrel \mid \rel,\lang), \p^*(\rrel \mid \rel,\lang))\raisetag{8pt}\label{eq:eval}
\end{align}
where
\begin{itemize}[noitemsep]
\item $\loss_\varepsilon(\hat{\p},\p^*)\equiv\max(|\hat{\p}-\p^*| - \varepsilon, 0)$
\item $\p^*(\rrel \mid \rel,\lang)=\frac{\cnt_{\lang}(\rrrel)}{\cnt_{\lang}(\rel)}$ is the empirical estimate of $\p(\rrel \mid \rel,\lang)$.
\item $\hat{\p}(\rrel \mid \rel,\lang)$ is the system's prediction of $p^*$
\end{itemize}
The aggregate metric \eqref{eq:eval} is an expected loss that is weighted by $\p^*(\rel \mid \lang) = \frac{\cnt_{\lang}(\rel)}{\sum_{\rel' \in\udrel}\cnt_{\lang}(\rel')}$, to emphasize relation types that are more frequent in $L$.

Why this loss function?  We chose an L1-style loss, rather than L2 loss or log-loss, so that the aggregate metric is not dominated by outliers.  We took $\varepsilon > 0$ in order to forgive small errors: if some predicted directionality is already ``in the ballpark,'' we prefer to focus on getting other predictions right, rather than fine-tuning this one.  Our intuition is that errors $< \varepsilon$ in $\mb{\hat{p}}$'s elements will not greatly harm downstream tasks that analyze individual sentences, and might even be easy to correct by grammar reestimation (e.g., EM) that uses $\mb{\hat{p}}$ as a starting point.

In short, we have the intuition that if our predicted $\mb{\hat{\p}}$ achieves small $\loss_\varepsilon$ on the frequent relation types, then $\mb{\hat{\p}}$ will be helpful for downstream tasks, although testing that intuition is beyond the scope of this paper.  One could tune $\varepsilon$ on a downstream task.

\section{Simple ``Expected Count'' Baseline}
\label{sec:ec}
\newcommand{\ec}{EC}
Before launching into our full models, we warm up with a simple baseline heuristic called {\em expected count} (EC), which is reminiscent of Principles \& Parameters. The idea is that if \pos{ADJ}s tend to precede nearby \pos{NOUN}s in the sentences of language $\lang$, then \deprel{amod} probably tends to point leftward in $L$.  After all, the training languages show that when \pos{ADJ} and \pos{NOUN} are nearby, they are usually linked by \deprel{amod}.

Fleshing this out, EC estimates directionalities as
\begin{align}\label{eq:ecdir}
  \hat{\p}(\rrel \mid \rel, \lang) = \frac{\ecnt_{\lang}(\rrrel)}{\ecnt_{\lang}(\rrrel) + \ecnt_{\lang}(\rlrel)}
\end{align}
where we estimate the expected $\rlrel$ and $\rrrel$ counts by
\begin{align}
  \ecnt_{\lang}(\rrrel) &= \sum_{u \in \ul}\;\sum_{\substack{1 \leq i < j \leq |u| \\ j-i < w}} \p(\rrrel\mid u_i, u_j) \label{eq:ecountr} \\
  \ecnt_{\lang}(\rlrel) &= \sum_{u \in \ul}\;\sum_{\substack{1 \leq i < j \leq |u| \\ j-i < w}} \p(\rlrel\mid u_i, u_j) \label{eq:ecountl}
\end{align}
Here $u$ ranges over tag sequences (sentences) of $\ul$, and $w$ is a window size that characterizes ``nearby.''\footnote{In our experiment, we chose $w=8$ by cross-validation over $w=2,4,8,16,\infty$.}

  In other words, we ask: given that $u_i$ and $u_j$ are nearby tag tokens in the test corpus $\ul$, are they likely to be linked?  Formulas \eqref{eq:ecountr}--\eqref{eq:ecountl} count such a pair as a ``soft vote'' for $\rrrel$ if such pairs tended to be linked by $\rrrel$ in the treebanks of the training languages,\footnote{Thus, the EC heuristic examines the correlation between relations and tags in the training treebanks.  But our methods in the next section will follow the formalization of
    \S\ref{sec:task}: they do not examine a training treebank beyond its directionality vector $\mb{\p^*}$.}
 and a ``soft vote'' for $\rlrel$ if they tended to be linked by $\rlrel$.

\textbf{Training}:
For any two tag types $t,t'$ in the universal POS tagset $\tagset$, we simply use the training treebanks to get empirical
estimates of $\p(\cdot \mid t,t')$,
taking
\begin{align}\label{eq:ec}
  \p(\rrrel\mid t, t') = \frac{\sum_{\lang} s_{\lang}\cdot\cnt_{\lang}(t \rrrel t')}{\sum_{\lang} s_{\lang}\cdot\cnt_{\lang}(t, t')}
\end{align}
and similarly for $\p(\rlrel\mid t, t')$.
This can be interpreted as the (unsmoothed) fraction of $(t, t')$ within a $w$-word window where $t$ is the $\rel$-type parent of $t'$, computed by micro-averaging over languages.  To get a fair average over languages, equation \eqref{eq:ec} downweights the languages that have larger treebanks, yielding a {\em weighted} micro-average in which we define the weight
$s_{\lang} = 1/\sum_{t \in \tagset, t'\in\tagset} \cnt_{\lang}(t,t')$.

As we report later in Table~\ref{tb:gi}, even this simple supervised heuristic performs significantly better than state-of-the-art grammar induction systems.  However, it is not a {\em trained} heuristic: it has no free parameters that we can tune to optimize our evaluation metric.  For example, it can pay too much attention to tag pairs that are not discriminative.  We therefore proceed to build a trainable, feature-based system.

\section{Proposed Model Architecture}\label{sec:architecture}

To train our model, we will try to minimize the evaluation objective \eqref{eq:eval} averaged over the training languages, plus a regularization term given in \S\ref{sec:training}.\footnote{We gave all training languages the same weight. In principle, we could have downweighted the synthetic languages as out-of-domain, using cross-validation to tune the weighting.}

\subsection{Directionality predictions from scores}

Our predicted directionality for relation $r$ will be
\begin{align}
  \hat{\p}(\rrel \mid \rel, \lang) = 1 / (1 + \exp(-\vs(\ul)_{\rel}))\label{eq:prob}
\end{align}
$\vs(\ul)$ is a parametric function (see \S\ref{sec:scoring} below) that maps $\ul$ to a \defn{score} vector in
$\mathbb{R}^\nudrel$.
Relation type $r$ should get positive or negative score according to whether it usually points right or left.
The formula above converts each score to a directionality---a probability in $(0,1)$---using a logistic transform.

\medskip

\subsection{Design of the scoring function \texorpdfstring{$\vs(\ul)$}{mlp}}\label{sec:scoring}

\begin{figure}[t]
\centering
\includegraphics[width=\columnwidth]{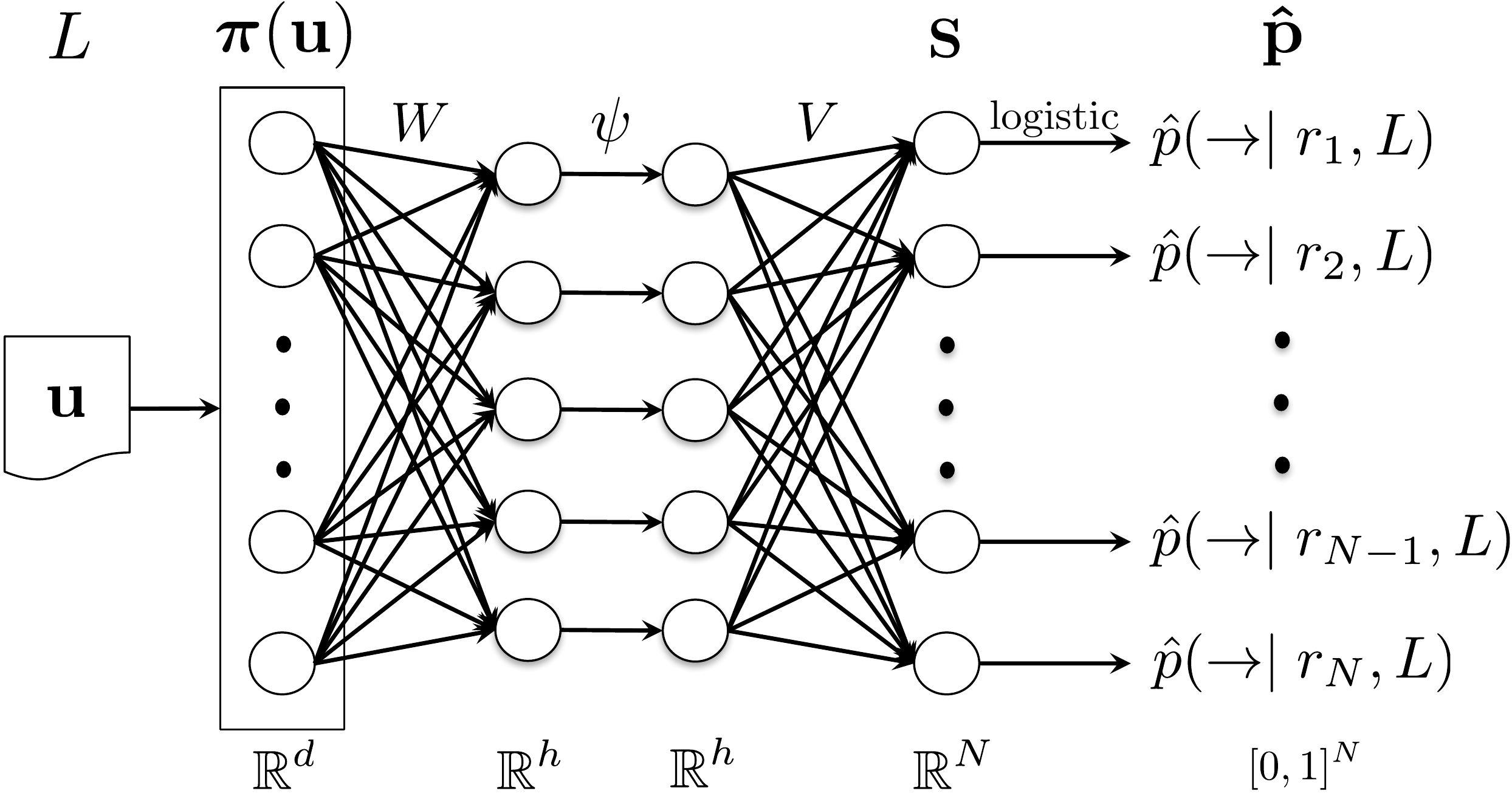}
\caption{\label{fig:mlp} Basic predictive architecture from equations~\eqref{eq:prob}--\eqref{eq:score}. $\vec{b}_W$ and $\vec{b}_V$ are suppressed for readability. }
\end{figure}
To score all dependency relation types given the corpus $\ul$, we use a feed-forward neural network with one hidden layer (Figure \ref{fig:mlp}):
\begin{align}
  \vs(\ul) = V\,\psi(W\vec{\pi}(\ul) + \vec{b}_W)+\vec{b}_V\label{eq:score}
\end{align}
$\vec{\pi}(\ul)$ extracts a $d$-dimensional feature vector from the corpus $\ul$ (see \S\ref{sec:extract} below). $W$ is a $h\times d$ matrix that maps $\vec{\pi}(\ul)$ into a $h$-dimensional space and $\vec{b}_W$ is a $h$-dimensional bias vector. $\psi$ is an element-wise activation function. $V$ is a $\nudrel\times h$ matrix whose rows can be regarded as learned embeddings of the dependency relation types. $\vec{b}_V$ is a $\nudrel$-dimensional bias vector that determines the default rightwardness of each relation type.  We give details in \S\ref{sec:hyperparams}.

The hidden layer $\psi(W\vec{\pi}(\ul) + \vec{b}_W)$ can be regarded as a latent representation of the language's word order properties, from which potentially {\em correlated} predictions $\mb{\hat{\p}}$ are extracted.

\subsection{Design of the featurization function \texorpdfstring{$\vec{\pi}(\ul)$}{feature}}\label{sec:extract}

Our current feature vector $\vec{\pi}(\ul)$ considers only the POS tag sequences for the sentences in the unparsed corpus $\ul$.  Each sentence is augmented with a special boundary tag \bdry at the start and end.  We explore both hand-engineered features and neural features.

\paragraph{Hand-engineered features.}  Recall that \S\ref{sec:ec} considered which tags appeared near one another in a given order.  We now devise a slew of features to measure such co-occurrences in a variety of ways.  By training the weights of these many features, our system will discover which ones are actually predictive.

Let $g(t \mid j) \in [0,1]$ be some measure (to be defined shortly) of the \defn{prevalence} of tag $t$ near token $j$ of corpus $\ul$.  We can then measure the prevalence of $t$, both overall and just near tokens of tag $s$:\footnote{\label{foot:backoff}In practice, we do backoff smoothing of these means.  This avoids
  subsequent division-by-0 errors if tag $t$ or $s$ has count 0 in the corpus, and it regularizes $\pi_{t|s}/\pi_{t}$ toward 1 if $t$ or $s$ is rare.  Specifically, we augment the denominators by adding $\lambda$, while augmenting the numerator in \eqref{eq:uni-prevalence} by adding $\lambda \cdot \mean_{j,t} g(t \mid j)$ (unsmoothed) and the numerator in \eqref{eq:bi-prevalence} by adding $\lambda$ times the smoothed $\pi_t$ from \eqref{eq:uni-prevalence}.  $\lambda > 0$ is a hyperparameter (see \S\ref{sec:hyperparams}).}
\begin{align}
\pi_{t} &= \mean_{j} g(t \mid j) \label{eq:uni-prevalence} \\
\pi_{t|s} &= \mean_{j:\, T_j=s} g(t \mid j) \label{eq:bi-prevalence}
\end{align}
where $T_j$ denotes the tag of token $j$.  We now define versions of these quantities for particular prevalence measures $g$.

Given $w > 0$, let the \defn{right window} $W_j$ denote the sequence of tags $T_{j+1},\ldots,T_{j+w}$ (padding this sequence with additional \bdry symbols if it runs past the end of $j$'s sentence).  We define quantities $\pi_{t|s}^w$ and $\pi_{t}^w$ via \eqref{eq:uni-prevalence}--\eqref{eq:bi-prevalence}, using a version of $g(t \mid j)$ that measures the fraction of tags in $W_j$ that equal $t$.  Also, for $b \in \{1,2\}$, we define $\pi_{t|s}^{w,b}$ and $\pi_{t}^{w,b}$ using a version of $g(t \mid j)$ that is 1 if $W_j$ contains at least $b$ tokens of $t$, and 0 otherwise.

For each of these quantities, we also define a corresponding \defn{mirror-image} quantity (denoted by negating $w > 0$) by computing the same feature on a reversed version of the corpus.

We also define ``truncated'' versions of all quantities above, denoted by writing $\hat{\phantom{w}}$ over the $w$.  In these, we use a \defn{truncated window} $\hat{W}_j$, obtained from $W_j$ by removing any suffix that starts with \bdry or with a copy of tag $T_j$ (that is, $s$).\footnote{In the ``fraction of tags'' features, $g(t \mid j)$ is undefined ($\frac{0}{0}$) when $\hat{W}_j$ is empty.  We omit undefined values from the means.}  As an example, $\pi_{\texttt{N}|\texttt{V}}^{\hat{8},2}$ asks how often a verb is followed by at least 2 nouns, within the next 8 words of the sentence {\em and before the next verb}.  A high value of this is a plausible indicator of a VSO-type or VOS-type language.

We include the following features for each tag pair $s,t$ and each $w \in \{1,3,8,100,$ $-1,-3,-8,-100,\hat{1},\hat{3},\hat{8},\hat{100},-\hat{1},-\hat{3},-\hat{8},-\hat{100}\}$:\footnote{The reason we don't include $\pi_{t|s}^{-w}//\pi_{t|s}^w$ is that it is included when computing features for $-w$.}
\begin{align*}
  \pi_t^w,\; \pi_{t|s}^w,\; \pi_{t|s}^w\cdot \pi_s^w,\; \pi_{t|s}^w//\pi_t^w,\; \pi_t^w//\pi_{t|s}^w,\; \pi_{t|s}^w//\pi_{t|s}^{-w}
\end{align*}
where we define $x//y = \min(x/y,1)$ to prevent unbounded feature values, which can result in poor generalization.  Notice that for $w=1$,
$\pi_{t|s}^w$ is bigram conditional probability, $\pi_{t|s}^w\cdot\pi_s^w$ is bigram joint probability, the log of $\pi_{t|s}^w/\pi_t^w$ is bigram pointwise mutual information, and $\pi_{t|s}^w/\pi_{t|s}^{-w}$ measures how much more prevalent $t$ is to the right of $s$ than to the left.
By also allowing other values of $w$, we generalize these features.
Finally, our model also uses versions of these features for each $b \in {1,2}$.

\begin{figure}[t]
\centering
\includegraphics[width=0.95\columnwidth]{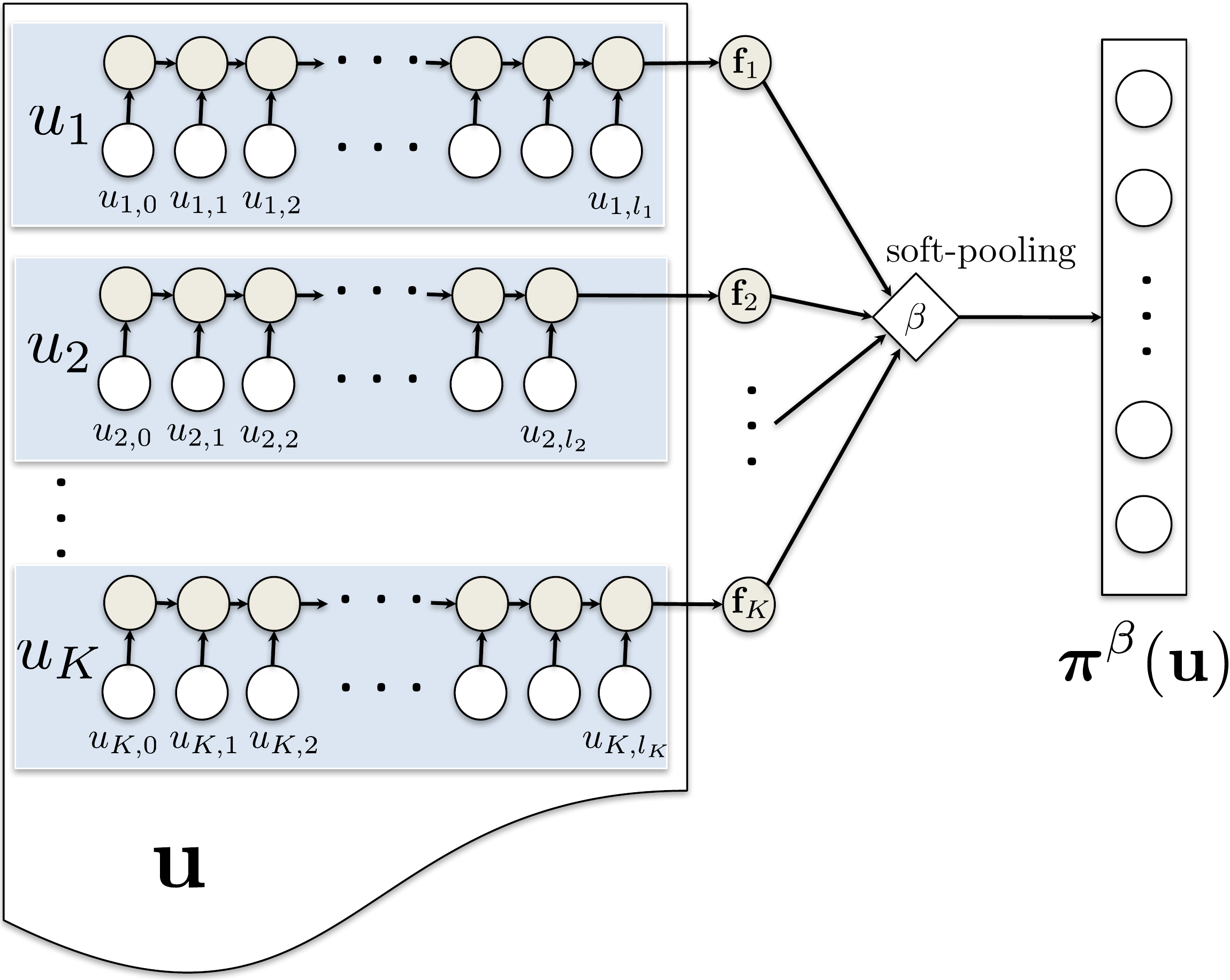}
\caption{\label{fig:nn}Extracting and pooling the neural features.}
\end{figure}

\paragraph{Neural features.} As an alternative to the manually designed $\vec{\pi}$ function above, we consider a neural approach to detect predictive configurations in the sentences of $\ul$, potentially including complex long-distance configurations.
Linguists working with Principles \& Parameters theory have supposed that a single telltale sentence---a \defn{trigger}---may be enough to determine a typological parameter, at least given the settings of other parameters \cite{gibson-wexler-1994,frank-kapur-1996}.

We map each corpus sentence $u_i$ to a finite-dimensional real vector $\vec{f}_i$ by using a gated recurrent unit (GRU) network \cite{DBLP:journals/corr/ChoMBB14}, a type of recurrent neural network that is a simplified variant of an LSTM network \cite{LSTM-1997}.  The GRU reads the sequence of one-hot embeddings of the tags in $u_i$ (including the boundary symbols \bdry).  We omit the part of the GRU that computes an output sequence, simply taking $\vec{f}_i$ to be the final hidden state vector.  The parameters are trained jointly with the rest of our typology prediction system, so the training procedure attempts to discover predictively useful configurations.

The various elements of $\vec{f}_i$ attempt to detect various interesting configurations in sentence $u_i$.  Some might be triggers (which call for max-pooling over sentences); others might provide softer evidence (which calls for mean-pooling).  For generality, therefore, we define our feature vector $\vec{\pi}(\ul)$ by \defn{soft-pooling} of the sentence vectors $\vec{f}_i$ (Figure \ref{fig:nn}).  The $\tanh$ gate in the GRU implies $f_{ik} \in (-1,1)$ and we transform this to the positive quantity $f'_{ik}=\frac{f_{ik}+1}{2} \in (0,1)$.  Given an ``inverse temperature'' $\beta$, define\footnote{For efficiency, we restrict the mean to $i \leq 1e4$ (the first 10,000 sentences).}
\begin{align}
  \pi^{\beta}_k = \left( \mean_i\;(f'_{ik})^{\beta} \right)^{1/\beta}
\end{align}
This $\pi^{\beta}_k$ is a pooled version of $f'_{ik}$, ranging from max-pooling as $\beta \rightarrow -\infty$ (i.e., does $f'_{ik}$ fire strongly on any sentence $i$?) to min-pooling as $\beta \rightarrow -\infty$.  It passes through arithmetic mean at $\beta=1$ (i.e., how strongly does $f'_{ik}$ fire on the average sentence $i$?), geometric mean as $\beta \rightarrow 0$ (this may be regarded as an arithmetic mean in log space), and harmonic mean at $\beta=-1$ (an arithmetic mean in reciprocal space).\looseness=-1

Our final $\vec{\pi}$ is a concatenation of the $\vec{\pi}^\beta$ vectors for $\beta \in \{-4,-2,-1,0,1,2,4\}$.  We chose these $\beta$ values experimentally, using cross-validation.

\paragraph{Combined model.}
We also consider a model
\begin{align}
  \vs(\ul) = \alpha\,\vs_{\text{H}}(\ul)+(1-\alpha)\,\vs_{\text{N}}(\ul)\label{eq:comb}
\end{align}
where $\vs_{\text{H}}(\ul)$ is the score assigned by the hand-feature system, $\vs_{\text{N}}(\ul)$ is the score assigned by the neural-feature system, and $\alpha \in [0,1]$ is a hyperparameter to balance the two.  $\vs_{\text{H}}(\ul)$ and $\vs_{\text{N}}(\ul)$ were trained separately.  At test time, we use \eqref{eq:comb} to combine them linearly before the logistic transform \eqref{eq:prob}.  This yields a weighted-product-of-experts model.

\subsection{Training procedure}\label{sec:training}

\paragraph{Length thresholding.} By default, our feature vector $\vec{\pi}(\ul)$ is extracted from those sentences in $\ul$ with length $\leq 40$ tokens.  In \S\ref{sec:subsets}, however, we try concatenating this feature vector with one that is extracted in the same way from just sentences with length $\leq 10$.
The intuition \cite{spitkovsky_baby_2010} is that the basic word order of the language can be most easily discerned from short, simple sentences.

\paragraph{Initialization.} We initialize the model of \eqref{eq:prob}--\eqref{eq:score} so
 that the estimated directionality $\hat{p}(\rrel \mid \rel, \lang)$, regardless of $\lang$, is initially a weighted mean of $\rel$'s directionalities in the training languages, namely
\begin{align}
\bar{p}_\rel &\equiv \sum_{\lang} \weight_{\lang}(\rel)\,p^*(\rrel \mid \rel,\lang)\label{eq:pbar} \\
\text{where }\weight_{\lang}(\rel) &\equiv \tfrac{p^*(\rel \mid \lang)}{\sum_{\lang'} p^*(\rel \mid \lang')}\label{eq:relweight}
\end{align}

\noindent This is done by setting $V=0$ and the bias $(\vec{b}_V)_\rel = \log \frac{\bar{p}_\rel}{1-\bar{p}_\rel}$, clipped to the range $[-10,10]$.
  
  As a result, we make sensible initial predictions even for rare relations $\rel$, which allows us to converge reasonably quickly even though we do not update the parameters for rare relations as often.

We initialize the recurrent connections in the GRU to random orthogonal matrices.  All other weight matrices in Figure~\ref{fig:mlp} and the GRU use ``Xavier initialization'' \cite{Glorot10understandingthe}. All other bias weight vectors are initialized to 0.

\paragraph{Regularization.}  We add an L2 regularizer to the objective.  When training the neural network, we use dropout as well.  All hyperparameters (regularization coefficient, dropout rate, etc.) are tuned via cross-validation; see \S\ref{sec:hyperparams}.

\paragraph{Optimization.} We use different algorithms in different feature settings. With scoring functions that use only hand features, we adjust the feature weights by stochastic gradient descent (SGD).  With scoring functions that include neural features, we use \mbox{RMSProp} \cite{Tieleman2012}.

\begin{table}
  \centering
  \begin{small}
    \begin{tabular}{|cc|c|}
      \hline
      \multicolumn{2}{|c|}{Train}&Test\\
      \hline
\parbox[t]{2cm}{\raggedright cs, es, fr, hi, de, it, la\_itt, no, ar, pt}&\parbox[t]{2.1cm}{\raggedright en, nl, da, fi, got, grc, et, la\_proiel, grc\_proiel, bg}&\parbox[t]{2.4cm}{\raggedright la, hr, ga, he, hu, fa, ta, cu, el, ro, sl, ja\_ktc, sv, fi\_ftb, id, eu, pl}\\
      \hline
    \end{tabular}
  \end{small}
  \caption{\label{tb:split} Data split of the 37 real languages, adapted from \newcite{wang-eisner-2016}.  (Our ``Train,'' on which we do 5-fold cross-validation, contains both their ``Train'' and ``Dev'' languages.)}
\end{table}

\section{Experiments}\label{sec:experiment}

\subsection{Data splits}\label{sec:splits}

We hold out 17 UD languages for testing (Table \ref{tb:split}).  For training, we use the remaining 20 UD languages and tune the hyperparameters with 5-fold cross-validation. That is, for each fold, we train the system on 16 real languages and evaluate on the remaining 4. When augmenting the 16 real languages with GD languages, we include only GD languages that are generated by ``mixing-and-matching'' those 16 languages, which means that we add $16\times 17\times 17=4624$ synthetic languages.\footnote{Why $16 \times 17 \times 17$?  As \newcite[\S5]{wang-eisner-2016} explain, each GD treebank is obtained from the UD treebank of some \defn{substrate} language $S$ by stochastically permuting the dependents of verbs and nouns to respect typical orders in the \defn{superstrate} languages $\RV$ and $\RN$ respectively.  There are 16 choices for $S$.  There are 17 choices for $\RV$ (respectively $\RN$), including $\RV=S$ (``self-permutation'') and $\RV=\emptyset$ (``no permutation'').}

Each GD treebank $\ul$ provides a standard split into train/dev/test portions.  In this paper, we primarily restrict ourselves to the train portions (saving the gold trees from the dev and test portions to tune and evaluate some future grammar induction system that consults our typological predictions).  For example, we write $\ul_{\text{train}}$ for the POS-tagged sentences in the ``train'' portion, and $\bm{\p^*}_{\text{train}}$ for the empirical probabilities derived from their gold trees.
We always train the model to predict $\bm{\p^*}_{\text{train}}$ from $\ul_{\text{train}}$ on each {\em training language}.
  To evaluate on a {\em held-out language} during cross-validation, we can measure how well the model predicts $\bm{\p^*}_{\text{train}}$ given $\ul_{\text{train}}$.\footnote{In actuality, we measured how well it predicts $\bm{\p^*}_{\text{dev}}$ given $\ul_{\text{dev}}$.  That was a slightly less sensible choice.  It may have harmed our choice of hyperparameters, since dev is smaller than train and therefore $\bm{\p^*}_{\text{dev}}$ tends to have greater sampling error.  Another concern is that our typology system, having been specifically tuned to predict $\bm{\p^*}_{\text{dev}}$, might provide an unrealistically accurate estimate of $\bm{\p^*}_{\text{dev}}$ to some future grammar induction system that is being cross-validated against the same dev set, harming that system's choice of hyperparameters as well.
}
For our final test, we evaluate on the 17 test languages using a model trained on all training languages (20 treebanks for UD, plus $20\times 21\times 21=8840$ when adding GD) with the chosen hyperparameters.  To evaluate on a {\em test language}, we again measure how well the model predicts $\bm{\p^*}_{\text{train}}$ from $\ul_{\text{train}}$.

\subsection{Comparison of architectures}\label{sec:experiment-architectures}

\begin{table}[t]
  \centering
\begin{small}
\begin{tabular}{r|c|c|c}
\multicolumn{2}{c|}{Architecture}&\multicolumn{2}{c}{$\varepsilon$-insensitive loss}\\\hline
Scoring&Depth&UD&+GD\\\hline
EC &-&0.104&0.099\\\hline
$\vs_{\text{H}}$ &0&0.057&\bf{0.037}*\\
$\vs_{\text{H}}$ &1&\bf{0.050}&\bf{0.036}*\\
$\vs_{\text{H}}$ &3&0.060&0.048\\
$\vs_{\text{N}}$ &1&\bf{0.062}&0.048\\
$\alpha\,\vs_{\text{H}} + (1-\alpha)\,\vs_{\text{N}}$ &1&\bf{0.050}&\bf{0.032}*\\
\end{tabular}
\end{small}
\caption{\label{tb:architecture} Average expected loss over 20 UD languages, computed by 5-fold cross-validation.
  The first column indicates whether we score using hand-engineered features ($\vs_{\text{H}}$), neural features ($\vs_{\text{N}}$), or a combination (see end of \S\ref{sec:extract}).  As a baseline, the first line evaluates the EC (expected count) heuristic from \S\ref{sec:ec}. Within each {\em column}, we boldface the best (smallest) result as well as all results that are not significantly worse (paired permutation test by language, $p < 0.05$).
    A starred result is significantly better than the other model in the same {\em row}.}
\end{table}

Table~\ref{tb:architecture} shows the cross-validation losses (equation~\eqref{eq:eval}) that are achieved by different scoring architectures.
We compare the results when the model is trained on real languages (the ``UD'' column) versus on real languages plus synthetic languages (the ``+GD'' column).

The $\vs_{\text{H}}$ models here use a subset of the hand-engineered features, selected by the experiments in \S\ref{sec:subsets} below and corresponding to Table~\ref{tb:subsets} line~8.

Although Figure~\ref{fig:mlp} and equation~\eqref{eq:score} presented an ``depth-1'' scoring network with one hidden layer, Table~\ref{tb:architecture} also evaluates ``depth-$d$'' architectures with $d$ hidden layers.  The depth-0 architecture simply predicts each directionality separately using logistic regression (although our training objective is not the usual convex log-likelihood objective).

Some architectures are better than others.  We note that the hand-engineered features outperform the neural features---though not significantly, since they make complementary errors---and that combining them is best.  However, the biggest benefit comes from augmenting the training data with GD languages; this consistently helps more than changing the architecture.

\subsection{Contribution of different feature classes}\label{sec:subsets}

\begin{table}[t]
  \centering
\begin{small}
\begin{tabular}{l|ll|l}
ID&Features&Length&Loss (+GD)\\\hline
0&$\varnothing$&---&0.076\\
1&conditional&40&0.058\\
2&joint&40&0.057\\
3&PMI&40&0.039\\
4&asymmetry&40&0.041\\\hline
5&rows 3+4&40&0.038\\
6&row 5+b&40&\bf{0.037}\\
7&row 5+t&40&\bf{0.037}\\\hline
8*&row 5+b+t &40&\bf{0.036}\\\hline
9&row 8&10&0.043\\
10&row 8 &10+40&\bf{0.036}\\
\end{tabular}
\end{small}
\caption{\label{tb:subsets} Cross-validation losses with different subsets of hand-engineered features from \S\ref{sec:extract}.
``$\varnothing$'' is a baseline with no features (bias feature only), so it makes the same prediction for all languages.  ``conditional'' = $\pi_{t|s}^w$ features, ``joint'' = $\pi_{t|s}^w\cdot \pi_s^w$ features, ``PMI'' = $\pi_{t|s}^w//\pi_t^w$ and $\pi_t^w//\pi_{t|s}^w$ features, ``asymmetry'' = $\pi_{t|s}^w//\pi_{t|s}^{-w}$ features, ``b'' are the features superscripted by $b$, and ``t'' are the features with truncated window. ``+'' means concatenation.
The ``Length'' field refers to length thresholding (see \S\ref{sec:training}).  The system in the starred row is the one that we selected for row 2 of Table~\ref{tb:architecture}.
}
\end{table}

To understand the contribution of different hand-engineered features, we performed forward selection tests on the depth-1 system, including only some of the features.  In all cases, we trained in the ``+GD'' condition.  The results are shown in Table~\ref{tb:subsets}.  Any class of features is substantially better than baseline, but we observe that most of the total benefit can be obtained with just PMI or asymmetry features.  Those features indicate, for example, whether a verb tends to attract nouns to its right or left.  We did not see a gain from length thresholding.

\subsection{Robustness to noisy input}\label{sec:robust}
We also tested our directionality prediction system on noisy input (without retraining it on noisy input).  Specifically, we tested the depth-1 $\vs_{\text{H}}$ system.  This time, when evaluating on the dev split of a held-out language, we provided a noisy version of that input corpus that had been retagged by an automatic POS tagger \cite{nguyen-EtAl:2014:Demos}, which was trained on just 100 gold-tagged sentences from the train split of that language.  The average tagging accuracy over the 20 languages was only 77.26\%.  Nonetheless, the ``UD''-trained and ``+GD''-trained systems got respective losses of 0.052 and 0.041---nearly as good as in Table \ref{tb:architecture}, which used gold POS tags.

\subsection{Hyperparameter settings}\label{sec:hyperparams}

For each result in Tables~\ref{tb:architecture}--\ref{tb:subsets}, the hyperparameters were chosen by grid search on the cross-validation objective (and the table reports the best result).  For the remaining experiments, we select the depth-1 combined models \eqref{eq:comb} for both ``UD'' and ``+GD,'' as they are the best models according to Table~\ref{tb:architecture}.

The hyperparameters for the selected models are as follows: When training with ``UD,'' we took $\alpha=1$ (which ignores $\vs_{\text{N}}$), with hidden layer size $h=256$, $\psi=\sigmoid$, $\text{L2\_coeff}=0$ (no L2 regularization), and $\text{dropout}=0.2$.  When training with ``+GD,'' we took $\alpha=0.7$, with different hyperparameters for the two interpolated models: $\vs_{\text{H}}$ uses $h=128$, $\psi=\sigmoid$, $\text{L2\_coeff}=0$, and $\text{dropout}=0.4$, while $\vs_{\text{N}}$ uses $h=128$, $\text{emb\_size}=128$, $\text{rnn\_size}=32$, $\psi=\relu$, $\text{L2\_coeff}=0$, and $\text{dropout}=0.2$. For both ``UD'' and ``+GD'', we use $\lambda=1$ for the smoothing in footnote \ref{foot:backoff}.

\subsection{Comparison with grammar induction}\label{sec:gi}

\setlength{\textfloatsep}{10pt plus 1.0pt minus 2.0pt}
\setlength\tabcolsep{4pt}
\begin{table}[t]
\center
\begin{small}
\begin{tabular}{l|c|c||c|c||c|c}
    &MS13&N10&EC&$\varnothing$&UD&+GD\\\hline
loss&0.166&0.139&0.098&0.083&0.080&{\bf 0.039}\\
\end{tabular}
\end{small}
\caption{\label{tb:gi}Cross-validation average expected loss of the two grammar induction methods, MS13 \cite{marecek2013stop} and N10 \cite{naseem_using_2010}, compared to the EC heuristic of \S\ref{sec:ec} and our architecture of \S\ref{sec:architecture} (the version from the last line of Table~\ref{tb:architecture}).  In these experiments, the dependency relation types are ordered POS pairs.
N10 harnesses prior linguistic knowledge, but its improvement upon MS13 is not statistically  significant.  Both grammar induction systems are {\em significantly} worse than the rest of the systems, including even our two baseline systems, namely EC (the ``expected count'' heuristic from \S\ref{sec:ec}) and $\varnothing$ (the no-feature baseline system from Table~\ref{tb:subsets} line 0).  Like N10, these baselines make use of some cross-linguistic knowledge, which they extract in different ways from the training treebanks.
Among our own 4 systems, EC is significantly worse than all others, and +GD is significantly better than all others.  (Note: When training the {\em baselines}, we found that including the +GD languages---a bias-variance tradeoff--- harmed EC but helped $\varnothing$. The table reports the better result in each case.)}
\end{table}

Grammar induction is an alternative way to predict word order typology.  Given a corpus of a language, we can first use grammar induction to parse it into dependency trees, and then estimate the directionality of each dependency relation type based on these (approximate) trees.

However, what are the dependency relation types?  Current grammar induction systems produce unlabeled dependency edges.  Rather than try to obtain a UD label like $r=\deprel{amod}$ for each edge, we label the edge deterministically with a POS pair such as $r=(parent=\pos{NOUN}, child=\pos{ADJ})$.  Thus, we will attempt to predict the directionality of each POS-pair relation type.  For comparison, we retrain our supervised system to do the same thing.

For the grammar induction system, we try the implementation of DMV with stop-probability estimation by \newcite{marecek2013stop}, which is a common baseline for grammar induction \cite{le2015unsupervised} because it is language-independent, reasonably accurate, fast, and convenient to use. We also try the grammar induction system of \newcite{naseem_using_2010}, which is the state-of-the-art system on UD \cite{D16-1004}.  \newcite{naseem_using_2010}'s method, like ours, has prior knowledge of what typical human languages look like.

Table~\ref{tb:gi} shows the results.  Compared to \newcite{marecek2013stop}, \newcite{naseem_using_2010} gets only a small (insignificant) improvement---whereas our ``UD'' system halves the loss, and the ``+GD'' system halves it again.  Even our baseline systems are significantly more accurate than the grammar induction systems, showing the effectiveness of casting the problem as supervised prediction.

\subsection{Fine-grained analysis}

\begin{figure}[t]
\centering
\includegraphics[width=8cm, height=5.5cm]{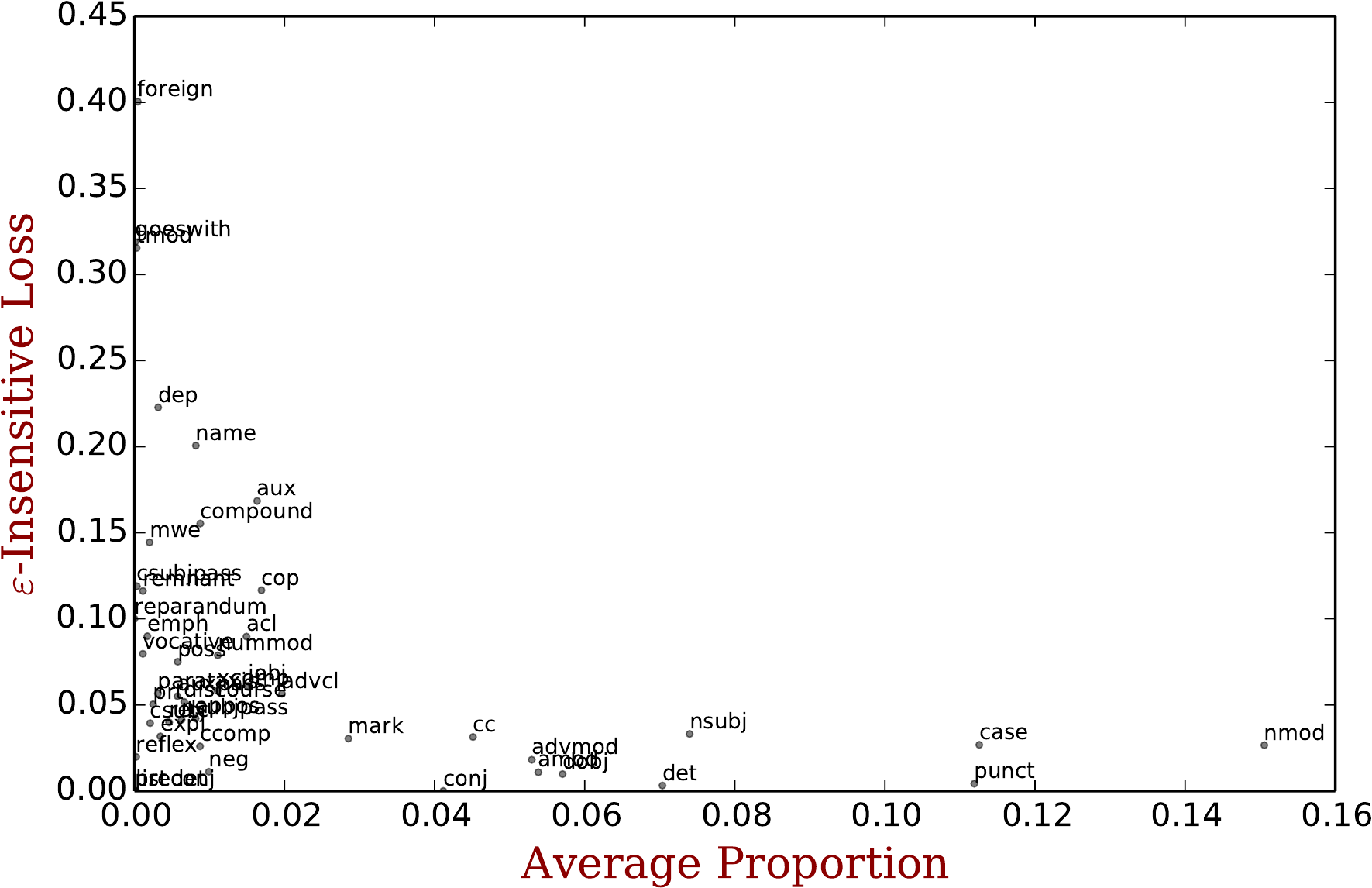}
\caption{\label{fig:ltype}Cross-validation loss broken down by relation.  We plot each relation $\rel$ with $x$ coordinate = the proportion of $\rel$ in the average training corpus = $\mean_{\lang \in \text{Train}} p^*_{\text{train}}(r \mid L) \in [0,1]$, and with $y$ coordinate = the weighted average \protect\linebreak $\sum_{\lang\in\text{Heldout}} \weight_{\lang}(\rel)\,\loss_{\varepsilon}(\hat{\p}_{\text{dev}}({\rrel} | \rel,\lang), \p^*_{\text{dev}}({\rrel} | \rel,\lang))$~(see~\eqref{eq:relweight}).}
\end{figure}

\begin{figure}[t]
\centering
\includegraphics[width=6cm, height=6cm]{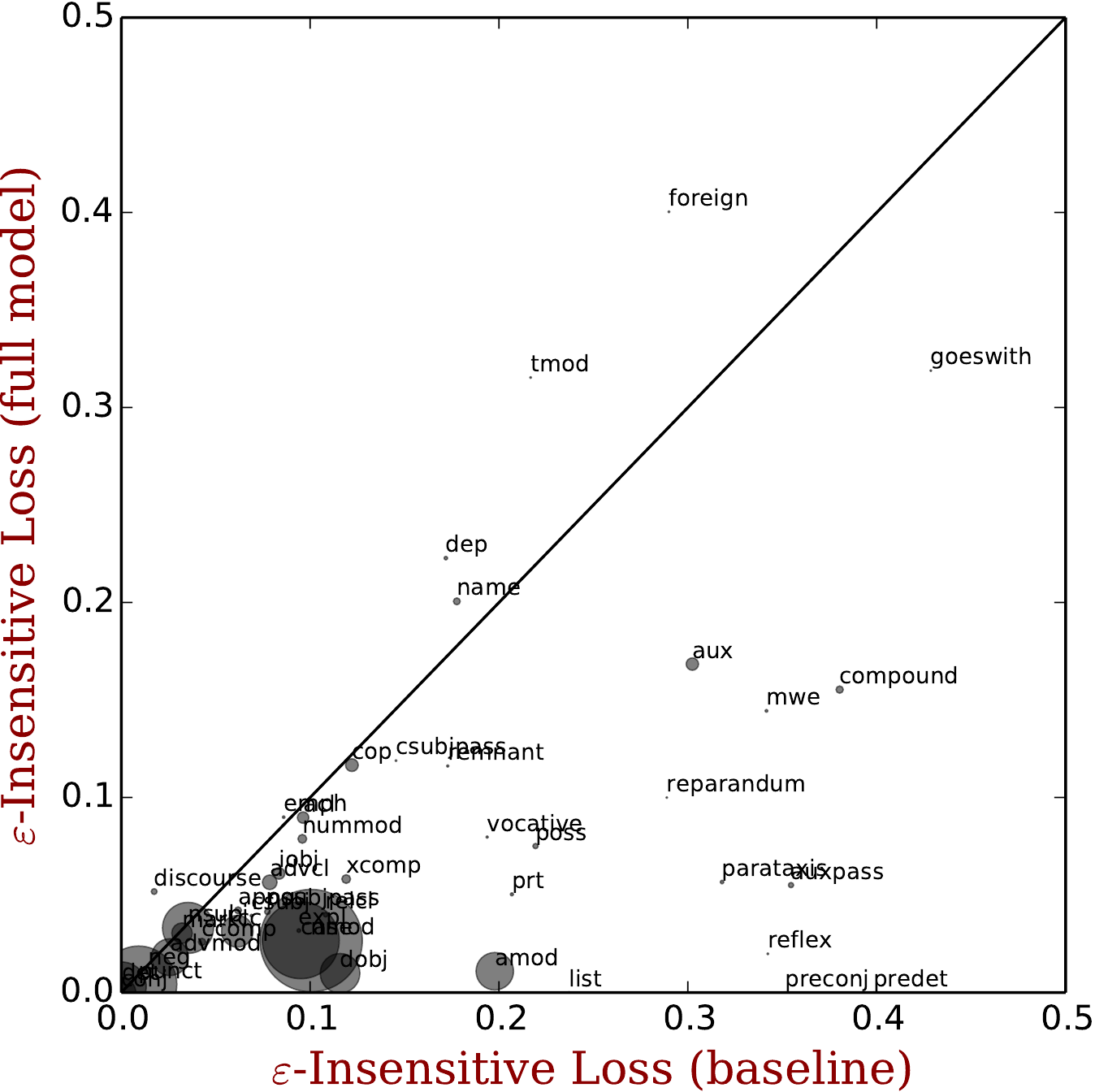}
\caption{\label{fig:ltype_pair}The $y$ coordinate is the average loss of our model (Table~\ref{tb:subsets} line 8), just as in Figure~\ref{fig:ltype}, whereas the $x$ coordinate is the average loss of a simple baseline model $\varnothing$ that ignores the input corpus (Table~\ref{tb:subsets} line 0).  Relations whose directionality varies more by language have higher baseline loss.  Relations that beat the baseline fall below the diagonal line.  The marker size for each relation is proportional to the $x$-axis in Figure~\ref{fig:ltype}.}
\end{figure}

Beyond reporting the aggregate cross-validation loss over the 20 training languages, we break down the cross-validation predictions by relation type.
Figure~\ref{fig:ltype} shows that the {\em frequent} relations are all quite predictable.
Figure~\ref{fig:ltype_pair} shows that our success is not just because the task is easy---on relations whose directionality varies by language, so that a baseline method does poorly, our system usually does well.

\begin{figure}[t]
\centering
\subfloat{\includegraphics[width=4cm, height=4cm]{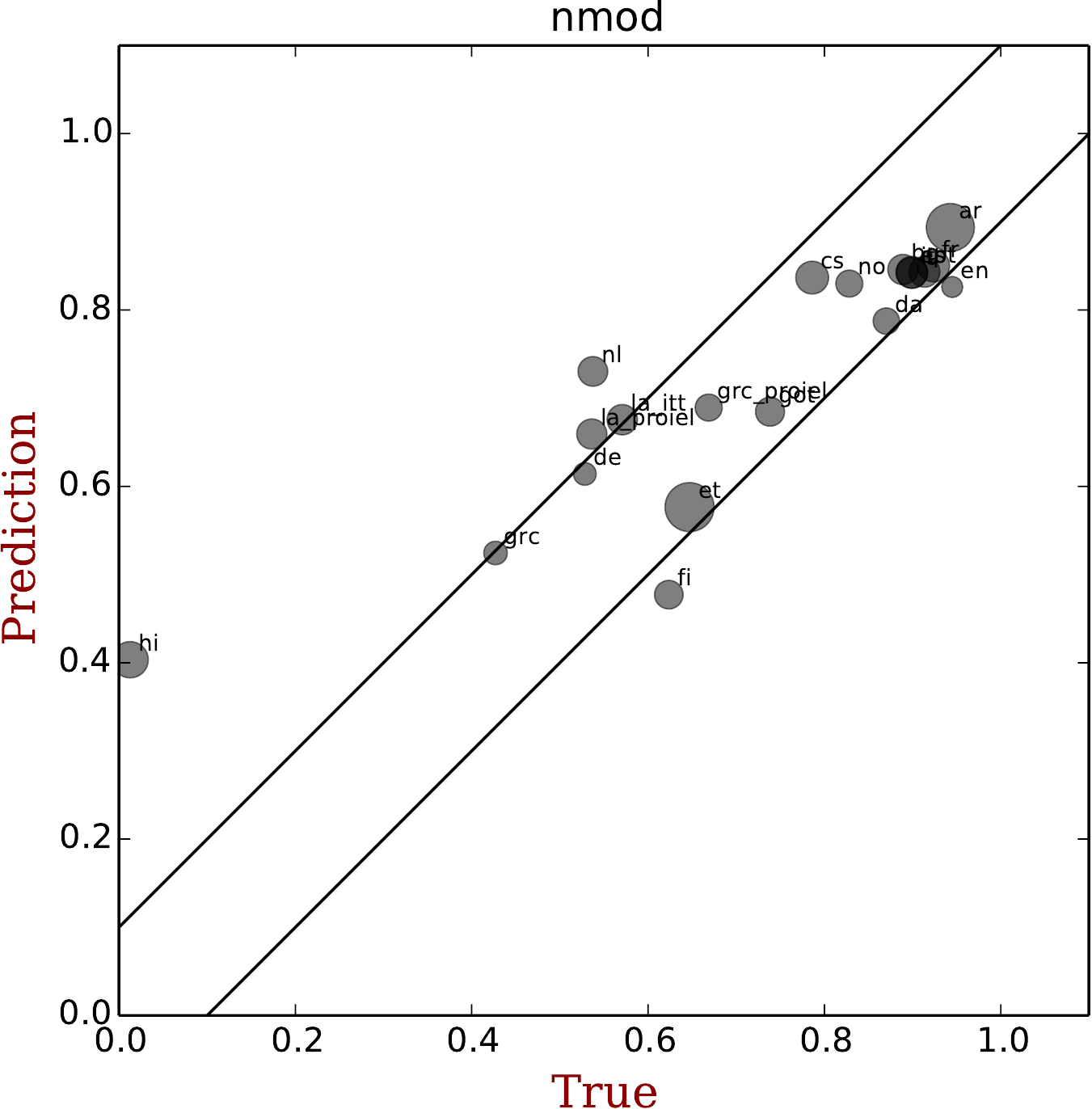}}
\subfloat{\includegraphics[width=4cm, height=4cm]{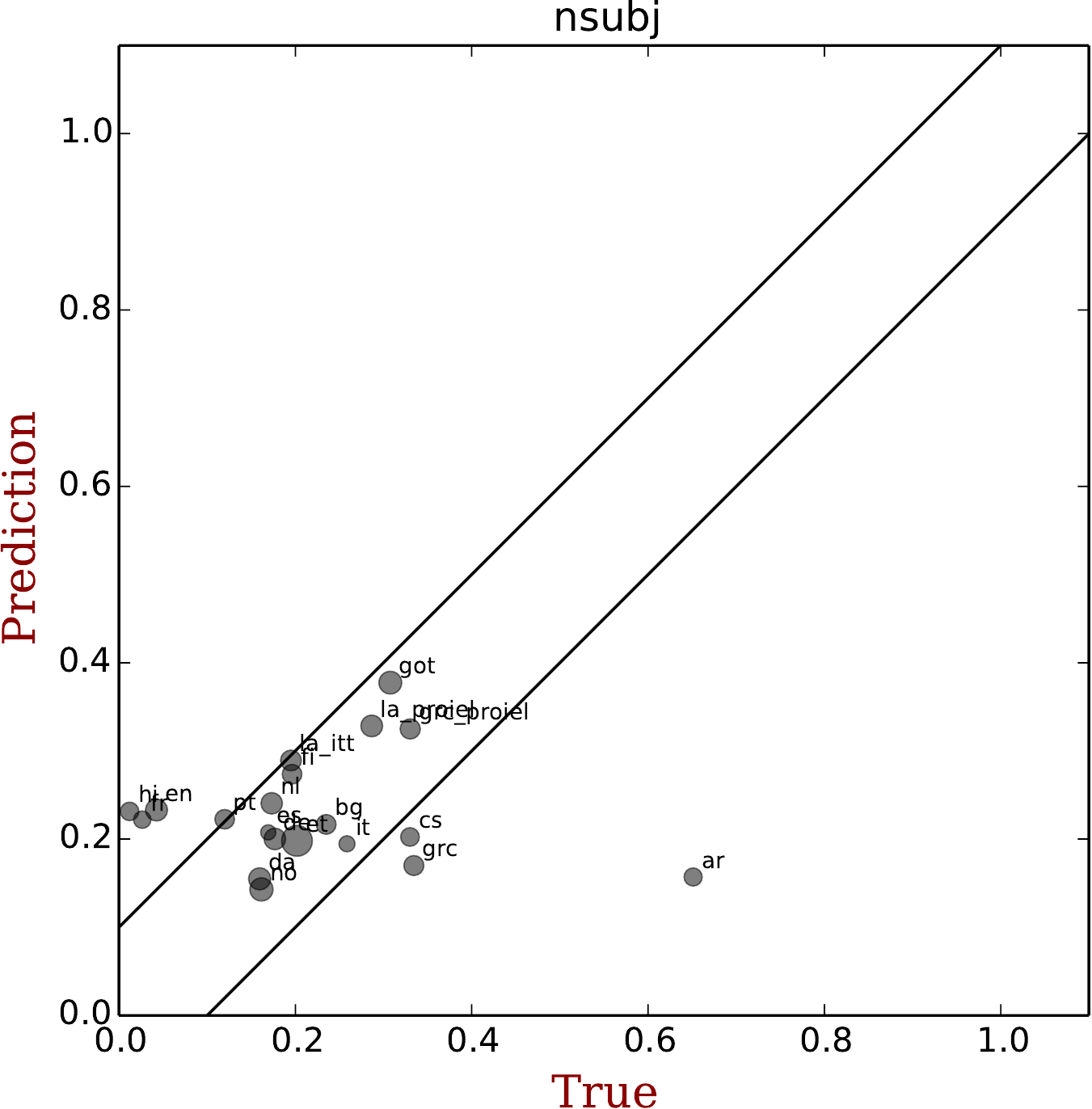}}\\[-3pt]
\subfloat{\includegraphics[width=4cm, height=4cm]{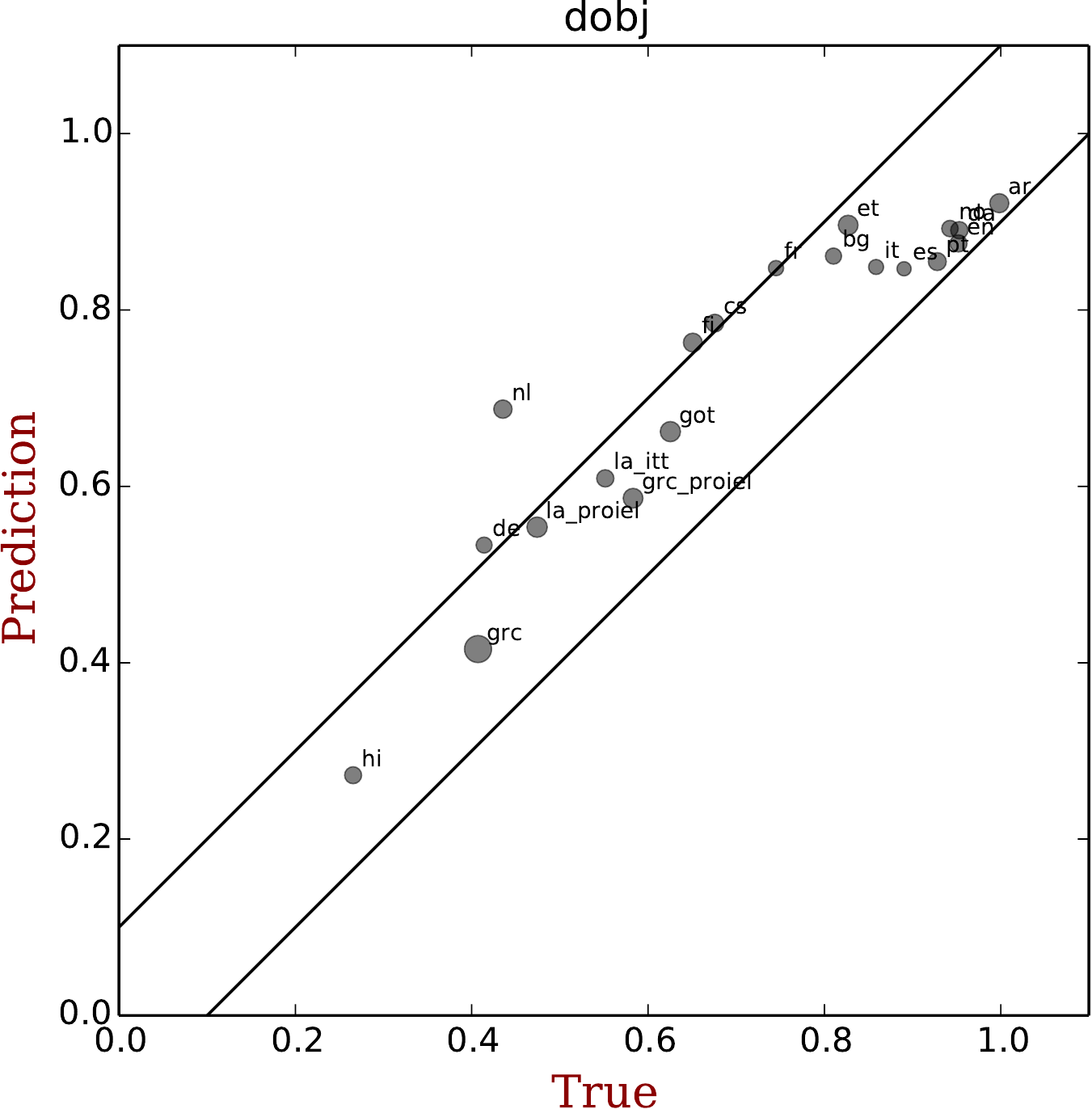}}
\subfloat{\includegraphics[width=4cm, height=4cm]{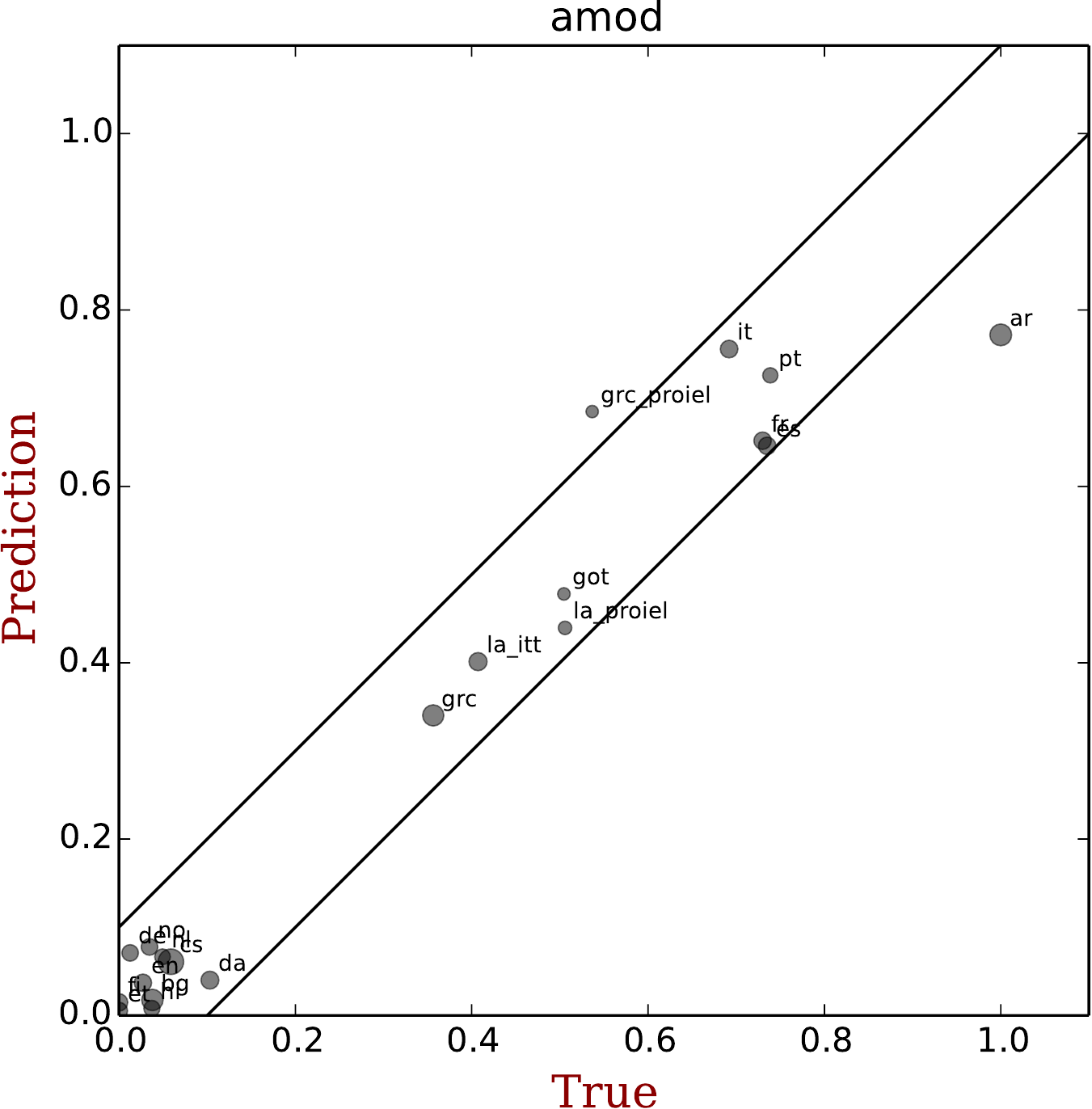}}\\[-3pt]
\subfloat{\includegraphics[width=4cm, height=4cm]{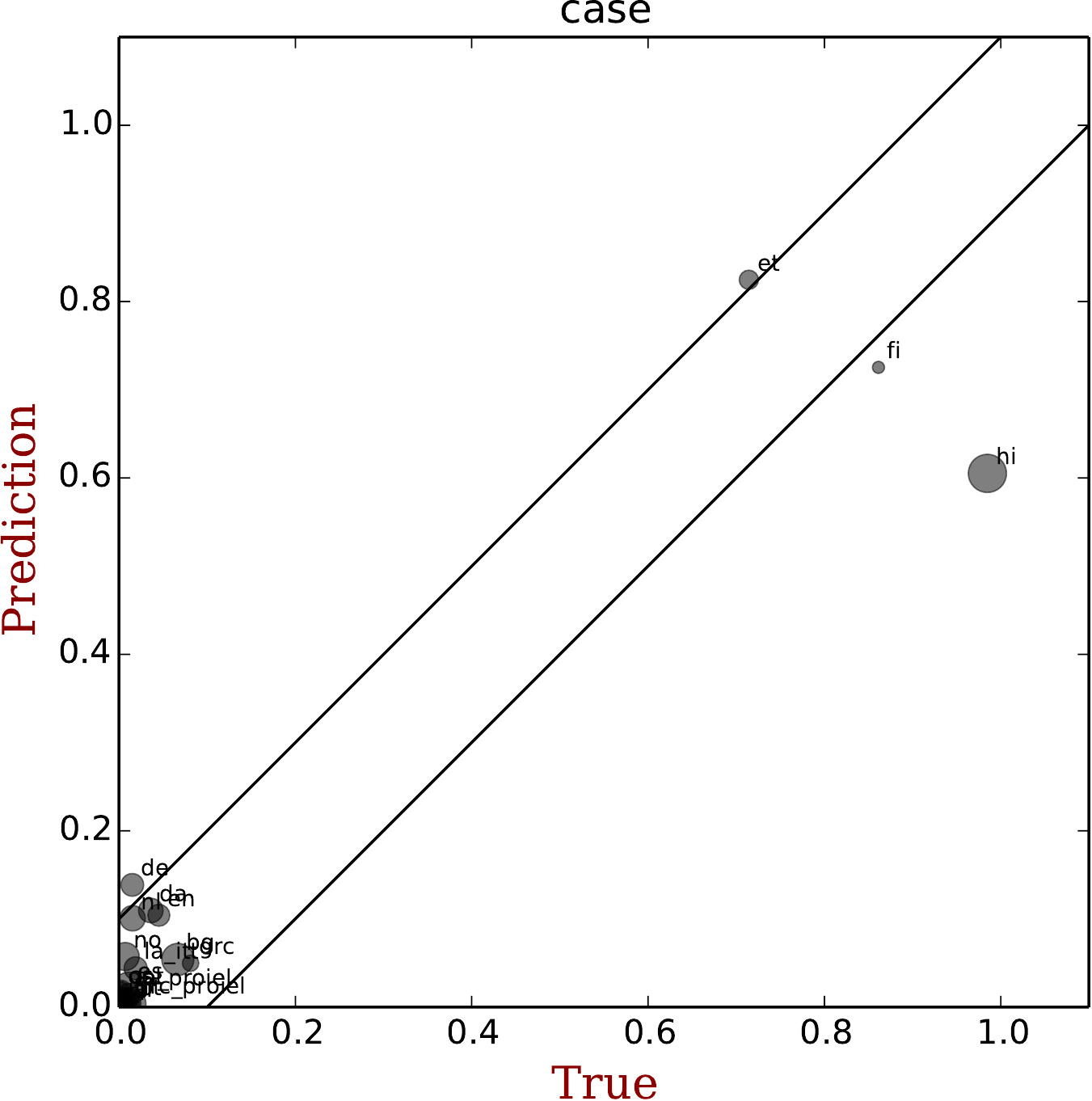}}
\subfloat{\includegraphics[width=4cm, height=4cm]{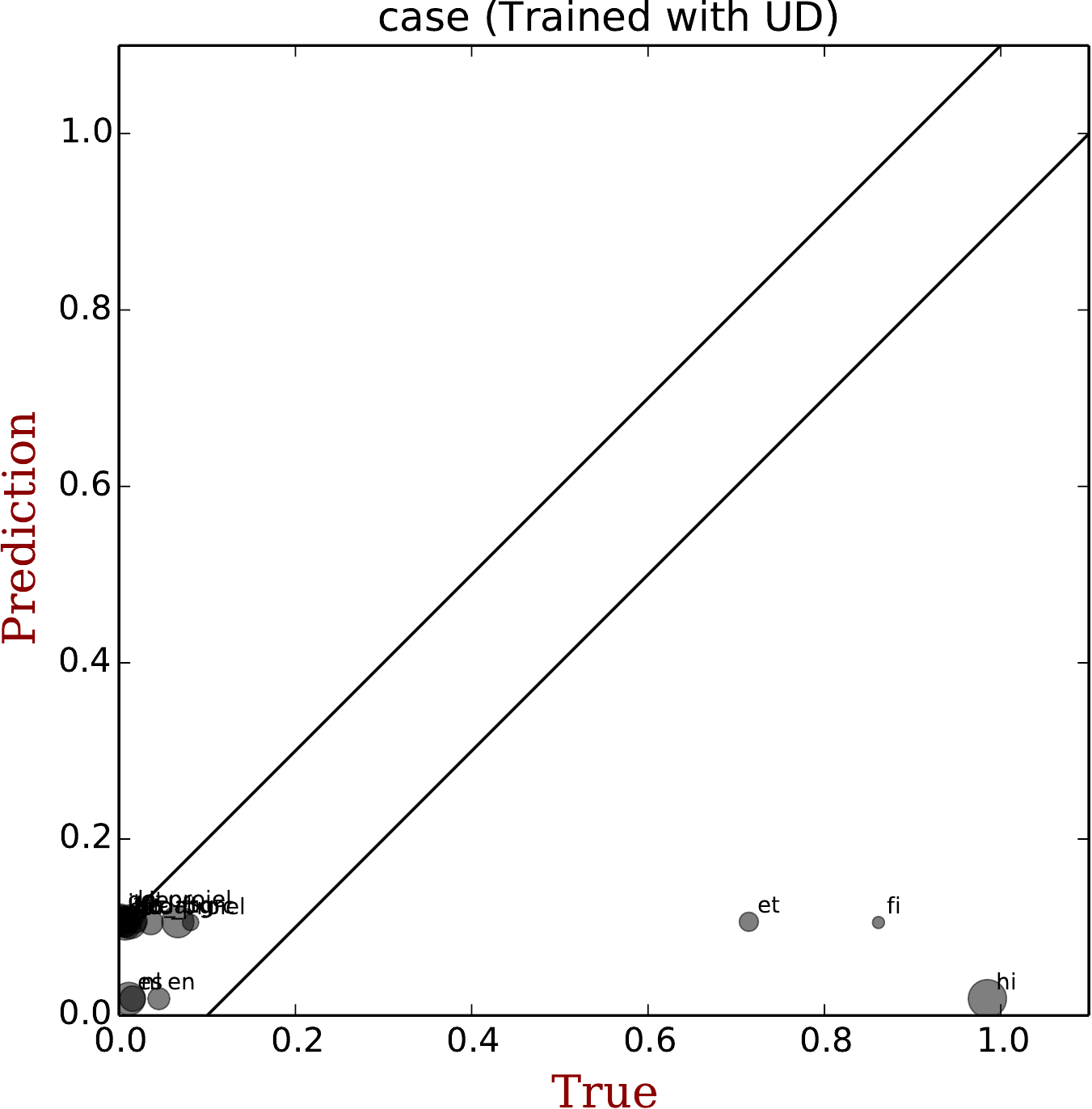}}
\caption{\label{fig:altype}Scatterplots of predicted vs.\@ true directionalities (by cross-validation). In the plot for relation type $\rel$, each language appears as a marker at $(\p^*, \hat{\p})$ (see \S\ref{sec:task}), with the marker size proportional to $\weight_{\lang}(\rel)$ (see \eqref{eq:relweight}). Points that fall between the solid lines ($|\hat{\p}-\p^*|\leq\varepsilon$) are considered ``correct,'' by the definition of $\varepsilon$-insensitive loss.  The last plot (bottom right) shows worse predictions for \deprel{case} when the model is trained on UD only.}
\end{figure}

To show that our system is behaving well across languages and not just on average, we zoom in on 5 relation types that are particularly common
or of particular interest to linguistic typologists.  These 5 relations together account for 46\% of all relation tokens in the average language:
\deprel{nmod} = noun-nominal modifier order,
\deprel{nsubj} = subject-verb order (feature 82A in the World Atlas of Language Structures),
\deprel{dobj} = object-verb order (83A),
\deprel{amod} = adjective-noun order (87A),
and \deprel{case} = placement of both adpositions and case markers (85A).

As shown in Figure \ref{fig:altype}, most points in the first five plots fall in or quite near the desired region.  We are pleased to see that the predictions are robust when the training data is unbalanced.  For example, the \deprel{case} relation points leftward in most real languages, yet our system can still predict the right directionality of ``hi'', ``et'' and ``fi.''  The credit goes to the diversity of our training set, which contains various synthetic \deprel{case}-right languages: the system fails on these three languages if we train on real languages only.  That said, apparently our training set is still not diverse enough to do well on the outlier ``ar'' (Arabic); see Figure 4 in \newcite{wang-eisner-2016}.

\subsection{Binary classification accuracy}
\label{sec:mva}
Besides $\varepsilon$-insensitive loss, we also measured how the systems perform on the coarser task of binary classification of relation direction.
We say that relation $\rel$ is dominantly ``rightward'' in language $\lang$ iff $\p^*(\rightarrow \mid \rel,\lang) > 0.5$.  We say that a system predicts ``rightward'' according to whether $\hat{\p}(\rightarrow \mid \rel,\lang) > 0.5$.

We evaluate whether this binary prediction is correct for each of the 20 most frequent relations $r$, for each held-out language $L$, using 5-fold cross-validation over the 20 training languages $L$ as in the previous experiment.
Tables \ref{tb:maj} and \ref{tb:majlang} respectively summarize these results by relation (equal average over languages) and by language (equal average over relations).  Keep in mind that these systems had not been specifically trained to place relations on the correct side of the artificial 0.5 boundary.

\label{sec:maj}
\begin{table}[t]
  \centering
\begin{footnotesize}
\begin{tabular}{cc@{\hspace{2ex}}|@{\hspace{2ex}}llll}
  Relation   &  Rate &   EC  &   $\varnothing$   &   UD  &   +GD \\\hline
  \deprel{  nmod    }   &   0.15    &   0.85    &   0.9 &   0.9 &   0.9 \\
  \deprel{  punct   }   &   0.11    &   0.85    &   0.85    &   0.85    &   0.85    \\
  \deprel{  case    }   &   0.11    &   0.75    &   0.85    &   0.85    &   1   \\
  \deprel{  nsubj   }   &   0.08    &   0.95    &   0.95    &   0.95    &   0.95    \\
  \deprel{  det }   &   0.07    &   0.8 &   0.9 &   0.9 &   0.9 \\
  \deprel{  dobj    }   &   0.06    &   0.6 &   0.75    &   0.75    &   0.85    \\
  \deprel{  amod    }   &   0.05    &   0.6 &   0.6 &   0.75    &   0.9 \\
  \deprel{  advmod  }   &   0.05    &   0.9 &   0.85    &   0.85    &   0.85    \\
  \deprel{  cc  }   &   0.04    &   0.95    &   0.95    &   0.95    &   0.95    \\
  \deprel{  conj    }   &   0.04    &   1   &   1   &   1   &   1   \\
  \deprel{  mark    }   &   0.03    &   0.95    &   0.95    &   0.95    &   0.95    \\
  \deprel{  advcl   }   &   0.02    &   0.85    &   0.85    &   0.85    &   0.8 \\
  \deprel{  cop }   &   0.02    &   0.75    &   0.75    &   0.65    &   0.75    \\
  \deprel{  aux }   &   0.02    &   0.9 &   0.6 &   0.75    &   0.65    \\
  \deprel{  iobj    }   &   0.02    &   0.45    &   0.55    &   0.5 &   0.6 \\
  \deprel{  acl }   &   0.01    &   0.45    &   0.85    &   0.85    &   0.8 \\
  \deprel{  nummod  }   &   0.01    &   0.9 &   0.9 &   0.9 &   0.9 \\
  \deprel{  xcomp   }   &   0.01    &   0.95    &   0.95    &   0.95    &   1   \\
  \deprel{  neg }   &   0.01    &   1   &   1   &   1   &   1   \\
  \deprel{  ccomp   }   &   0.01    &   0.75    &   0.95    &   0.95    &   0.95    \\\hline
    Avg.        &   -   &   0.81    &   0.8475  &   0.855   &   0.8775  \\
\hline\hline
\end{tabular}
\end{footnotesize}
\caption{\label{tb:maj}Accuracy on the simpler task of binary classification of relation directionality.
  The most common relations are shown first: the ``Rate'' column gives the average rate of the relation across the 20 training languages (like the $x$ coordinate in Fig.~\ref{fig:ltype}).}
\end{table}

Binary classification is an easier task.  It is easy because, as the $\varnothing$ column in Table~\ref{tb:maj} indicates, most relations have a clear directionality preference shared by most of the UD languages.  As a result, the better models with more features have less opportunity to help.  Nonetheless, they do perform better, and the EC heuristic continues to perform worse.

\begin{table}[t]
  \centering
\begin{footnotesize}
\begin{tabular}{c@{\hspace{2ex}}|@{\hspace{2ex}}llll}
  target    &   EC  &   $\varnothing$   &   UD  &   +GD \\\hline
  ar    &   0.8 &   0.8 &   0.75    &   0.85    \\
  bg    &   0.85    &   0.95    &   0.95    &   0.95    \\
  cs    &   0.9 &   1   &   1   &   0.95    \\
  da    &   0.8 &   0.95    &   0.95    &   0.95    \\
  de    &   0.9 &   0.9 &   0.9 &   0.95    \\
  en    &   0.9 &   1   &   1   &   1   \\
  es    &   0.9 &   0.9 &   0.95    &   0.95    \\
  et    &   0.8 &   0.8 &   0.8 &   0.8 \\
  fi    &   0.75    &   0.85    &   0.85    &   0.85    \\
  fr    &   0.9 &   0.9 &   0.9 &   0.95    \\
  got   &   0.75    &   0.8 &   0.85    &   0.8 \\
  grc   &   0.6 &   0.7 &   0.7 &   0.75    \\
  grc\_proiel   &   0.8 &   0.8 &   0.85    &   0.9 \\
  hi    &   0.6 &   0.45    &   0.45    &   0.7 \\
  it    &   0.9 &   0.9 &   0.9 &   0.95    \\
  la\_itt   &   0.7 &   0.85    &   0.8 &   0.85    \\
  la\_proiel    &   0.7 &   0.7 &   0.75    &   0.7 \\
  nl    &   0.95    &   0.85    &   0.85    &   0.85    \\
  no    &   0.9 &   1   &   1   &   0.95    \\
  pt    &   0.8 &   0.85    &   0.9 &   0.9 \\\hline
  Avg.  &   0.81    &   0.8475  &   0.855   &   0.8775  \\
\hline\hline
\end{tabular}
\end{footnotesize}
\vspace{-4pt}
\caption{\label{tb:majlang}
Accuracy on the simpler task of binary classification of relation directionality for each training language. A detailed comparison shows that EC is {\em significantly} worse than UD and +GD, and that $\varnothing$ is {\em significantly} worse than +GD (paired permutation test over the 20 languages, $p < 0.05$). The improvement from UD to +GD is {\em insignificant}, which suggests that this is an easier task where weak models might suffice.
}
\end{table}
In particular, EC fails significantly on \deprel{dobj} and \deprel{iobj}. This is because \deprel{nsubj}, \deprel{dobj}, and \deprel{iobj} often have different directionalities (e.g., in SVO languages), but the EC heuristic will tend to predict the same direction for all of them, according to whether \pos{NOUN}s tend to precede nearby \pos{VERB}s.

\subsection{Final evaluation on test data}\label{sec:testresults}

All previous experiments were conducted by cross-validation on the 20 training languages.  We now train the system on all 20, and report results on the 17 blind test languages (Table \ref{tb:finaltest}).  In our evaluation metric \eqref{eq:eval}, $\udrel$ includes all 57 relation types that appear in training data, plus a special {\sc unk} type for relations that appear only in test data.
The results range from good to excellent, with synthetic data providing consistent and often large improvements.

These results could potentially be boosted in the future by using an even larger and more diverse training set.  In principle, when evaluating on any one of our 37 real languages, one could train a system on {\em all} of the other 36 (plus the galactic languages derived from them), not just 20.  Moreover, the Universal Dependencies collection has continued to grow beyond the 37 languages used here (\S\ref{sec:data}).  Finally, our current setup extracts only one training example from each (real or synthetic) language.  One could easily generate a variant of this example each time the language is visited during stochastic optimization, by bootstrap-resampling its training corpus (to add ``natural'' variation) or subsampling it (to train the predictor to work on smaller corpora).  In the case of a synthetic language, one could also generate a corpus of new trees each time the language is visited (by re-running the stochastic permutation procedure, instead of reusing the particular permutation released by the Galactic Dependencies project).

\begin{table}[t]
  \centering
\begin{footnotesize}
\begin{tabular}{ccc|ccc}
\multicolumn{3}{c|}{Test}&\multicolumn{3}{c}{Train}\\\hline
target  &UD&+GD  &target   &UD&+GD\\\hline
cu&0.024&0.024  &   ar&0.116&0.057  \\
el&0.056&0.011  &   bg&0.037&0.015  \\
eu&0.250&0.072  &   cs&0.025&0.014  \\
fa&0.220&0.134  &   da&0.024&0.017  \\
fi\_ftb&0.073&0.029  &   de&0.046&0.025  \\
ga&0.181&0.154  &   en&0.025&0.036  \\
he&0.079&0.033  &   es&0.012&0.007  \\
hr&0.062&0.011  &   et&0.055&0.014  \\
hu&0.119&0.102  &   fi&0.069&0.070  \\
id&0.099&0.076  &   fr&0.024&0.018  \\
ja\_ktc&0.247&0.078  &   got&0.008&0.026 \\
la&0.036&0.004  &   grc&0.026&0.007 \\
pl&0.056&0.023  &   grc\_proiel&0.004&0.017  \\
ro&0.029&0.009  &   hi&0.363&0.191  \\
sl&0.015&0.031  &   it&0.011&0.008  \\
sv&0.012&0.008  &   la\_itt&0.033&0.023  \\
ta&0.238&0.053  &   la\_proiel&0.018&0.021   \\
&&  &   nl&0.069&0.066  \\
&&  &   no&0.008&0.010  \\
&&  &   pt&0.038&0.004  \\
\hline
$\!\!\!$Test Avg.&0.106&0.050*&All Avg.&0.076&0.040*\\\hline\hline
\end{tabular}
\end{footnotesize}
\caption{\label{tb:finaltest} Our final comparison on the 17 test languages appears at left.  We ask whether the average expected loss on these 17 real target languages is reduced by augmenting the training pool of 20 UD languages with +20*21*21 GD languages. For completeness, we extend the table with the cross-validation results on the training pool.  The ``Avg.'' lines report the average of 17 test or 37 training+testing languages. We mark both ``+GD'' averages with ``*'' as they are significantly better than their ``UD'' counterparts (paired permutation test by language, $p<0.05$).}
\end{table}

\section{Related Work}\label{sec:relatedwork}
\vspace{-2pt}

Typological properties can usefully boost the performance of cross-linguistic systems \cite{bender-2009,ohoran2016survey}. These systems mainly aim to annotate low-resource languages with help from models trained on similar high-resource languages. \newcite{naseem2012selective} introduce a ``selective sharing'' technique for generative parsing,
in which
a Subject-Verb language will use parameters shared with other Subject-Verb languages.  \newcite{tackstrom2013target} and \newcite{zhang-barzilay:2015:EMNLP} extend this idea to discriminative parsing and gain further improvements by conjoining regular parsing features with typological features. The cross-linguistic neural parser of \newcite{ammar2016one} conditions on typological features by supplying a ``language embedding'' as input.
\newcite{zhang-et-al-2012} use typological properties to convert language-specific POS tags to UD POS tags, based on their ordering in a corpus.

Moving from engineering to science, linguists seek {\em typological universals} of human language \cite{greenberg1963some,croft2002typology,song2014linguistic,hawkins2014word}, e.g.,
  ``languages with dominant Verb-Subject-Object order are always prepositional.'' \newcite{wals} characterize 2679 world languages with 192 typological properties. Their WALS database can supply features to NLP systems (see previous paragraph) or gold standard labels for typological classifiers.  \newcite{daume-campbell-2007} take WALS as input and propose a Bayesian approach to discover new universals.  \newcite{georgi-xia-lewis-2010} impute missing properties of a language, not by using universals, but by backing off to the language's typological cluster.  \newcite{murawaki2015continuous} use WALS
  to help recover the evolutionary tree of human languages; \newcite{daume-2009} considers the geographic distribution of WALS properties.\looseness=-1

Attempts at automatic typological classification are relatively recent.  \newcite{lewis-xia-2008} predict typological properties from induced trees, but guess those trees from aligned bitexts, not by monolingual grammar induction as in \S\ref{sec:gi}.
\newcite{liu-2010} and \newcite{futrell-mahowald-gibson-2015} show that the directionality of (gold) dependencies is indicative of ``basic'' word order and freeness of word order.  Those papers predict typological properties from trees that are automatically (noisily) annotated or manually (expensively) annotated.
An alternative is to predict the typology directly from raw or POS-tagged text, as we do. \newcite{roy-et-al-2014} first explored this idea, building a system that correctly predicts adposition typology on 19/23 languages with only word co-occurrence statistics. \newcite{zhang-et-al-2016} evaluate semi-supervised POS tagging by asking whether the induced tag sequences can predict typological properties.  Their prediction approach is supervised like ours, although developed separately and trained on different data.
They more simply predict 6 binary-valued WALS properties,
  using 6 independent binary classifiers based on POS bigram and trigrams.

Our task is rather close to grammar induction, which likewise predicts a set of real numbers giving the relative probabilities of competing syntactic configurations.  Most previous work on grammar induction begins with maximum likelihood estimation of some generative model---such as a PCFG \cite{lari1990estimation,Carroll92twoexperiments} or dependency grammar \cite{klein_corpus-based_2004}---though it may add linguistically-informed inductive bias \cite{ganchev2010posterior,naseem_using_2010}.
Most such methods use local search and must wrestle with local optima \cite{spitkovsky-alshawi-jurafsky-2013}.  Fine-grained typological classification might supplement this approach, by
cutting through the initial combinatorial challenge of establishing the basic word-order properties of the language.  In this paper we only quantify the directionality of each relation {\em type}, ignoring how {\em tokens} of these relations interact locally to give coherent parse trees.  Grammar induction methods like EM could naturally consider those local interactions for a more refined analysis, when guided by our predicted global directionalities.

\vspace{-3pt}
\section{Conclusions and Future Work}\label{sec:futurework}
\vspace{-3pt}

We introduced a typological classification task, which attempts to extract quantitative knowledge about a language's syntactic structure from its surface forms (POS tag sequences).  We applied supervised learning to this apparently unsupervised problem.  As far as we know, we are the first to utilize synthetic languages to train a learner for real languages: this move yielded substantial benefits.\footnote{Although \newcite{wang-eisner-2016} review uses of synthetic training data elsewhere in machine learning.}

Figure~\ref{fig:altype} shows that we rank held-out languages rather accurately along a spectrum of directionality, for several common dependency relations.  Table~\ref{tb:finaltest} shows that if we jointly predict the directionalities of {\em all} the relations in a new language, most of those numbers will be quite close to the truth (low aggregate error, weighted by relation frequency).  This holds promise for aiding grammar induction.

Our trained model is robust when applied to noisy POS tag sequences.  In the future, however, we would like to make similar predictions from raw word sequences.  That will require features that abstract away from the language-specific vocabulary.  Although recurrent neural networks in the present paper did not show a clear advantage over hand-engineered features, they might be useful when used with word embeddings.

Finally, we are interested in downstream uses.  Several NLP tasks have benefited from typological features (\S\ref{sec:relatedwork}).  By using end-to-end training, our methods could be tuned to extract features (existing or novel) that are particularly useful for some task.

\paragraph{Acknowledgements} This work was funded by the U.S. National Science Foundation under Grant No. 1423276.  We are grateful to the state of Maryland for providing indispensable computing resources via the Maryland Advanced Research Computing Center (MARCC). We thank the Argo lab members for useful discussions. Finally, we thank TACL action editor Mark Steedman and the anonymous reviewers for high-quality suggestions, including the EC baseline and the binary classification evaluation.
\bibliography{typology}
\bibliographystyle{tacl_plainnaturl}

\end{document}